\documentclass{article}

\usepackage[nonatbib,preprint]{neurips_2024}

\usepackage[utf8]{inputenc} 
\usepackage[T1]{fontenc}    
\usepackage{hyperref}       
\usepackage{url}            
\usepackage{booktabs}       
\usepackage{amsfonts}       
\usepackage{nicefrac}       
\usepackage{microtype}      
\usepackage{xcolor}         

\usepackage{amsmath}
\usepackage{graphicx}

\title{Improve Multi-Modal Embedding Learning via Explicit Hard Negative Gradient Amplifying}

\author{%
  Youze Xue\thanks{This work is a part of QQ MLLM project.}, Dian Li\thanks{Corresponding author. Project leader of QQ MLLM project.}, Gang Liu\\
  Tencent QQ \\
  \texttt{\{youzexue, goodli, sinbadliu\}@tencent.com} \\
}

\begin{document}

\maketitle

\begin{abstract}
  With the rapid advancement of multi-modal large language models (MLLMs) in recent years, the foundational Contrastive Language-Image Pretraining (CLIP) framework has been successfully extended to MLLMs, enabling more powerful and universal multi-modal embeddings for a wide range of retrieval tasks. Despite these developments, the core contrastive learning paradigm remains largely unchanged from CLIP-style models to MLLMs. Within this framework, the effective mining of hard negative samples continues to be a critical factor for enhancing performance. Prior works have introduced both offline and online strategies for hard negative mining to improve the efficiency of contrastive learning. While these approaches have led to improved multi-modal embeddings, the specific contribution of each hard negative sample to the learning process has not been thoroughly investigated.
  In this work, we conduct a detailed analysis of the gradients of the info-NCE loss with respect to the query, positive, and negative samples, elucidating the role of hard negatives in updating model parameters. Building upon this analysis, we propose to explicitly amplify the gradients associated with hard negative samples, thereby encouraging the model to learn more discriminative embeddings. Our multi-modal embedding model, trained with the proposed Explicit Gradient Amplifier and based on the LLaVA-OneVision-7B architecture, achieves state-of-the-art performance on the MMEB benchmark compared to previous methods utilizing the same MLLM backbone. Furthermore, when integrated with our self-developed MLLM, QQMM, our approach attains the top rank on the MMEB leaderboard.
  Code and models are released on \url{https://github.com/QQ-MM/QQMM-embed}.
  
\end{abstract}

\section{Introduction}

Multi-modal embedding models have found widespread application in areas such as multi-modal recommendation\cite{zhang2024notellm}\cite{zhang2024notellm2}, retrieval-augmented generation (RAG)\cite{yu2024visrag}\cite{tanaka2025vdocrag}, and multi-modal search engines\cite{jiangmmsearch}.
The Contrastive Language-Image Pretraining (CLIP) model established a standard framework in which text and image embeddings are aligned or separated based on their semantic relevance through contrastive learning\cite{CLIP}.
While CLIP-style models\cite{CLIP}\cite{SigLIP}\cite{SigLIP2}\cite{Eva-CLIP} have demonstrated effective cross-modal alignment between textual and visual modalities, their inherent design—processing unimodal inputs independently—limits their ability to generate unified embeddings for complex, universal multi-modal contexts.
Recent advances in multi-modal large language models (MLLMs)\cite{llava}\cite{llava-ov}\cite{qwen2-vl}\cite{InternVL} have demonstrated remarkable capabilities in understanding and reasoning over multi-modal inputs, motivating researchers to leverage pretrained MLLMs for enhanced embedding learning.
Owing to extensive pretraining and supervised fine-tuning (SFT) on large-scale datasets, MLLMs are capable of analyzing intricate semantic relationships between language and visual inputs.
Moreover, they exhibit flexibility in adapting to various retrieval tasks by utilizing different instruction prompts. Consequently, the focus of multi-modal embedding has shifted from merely extracting textual or visual features to generating comprehensive, instruction-guided representations that capture deep semantic information across modalities.

In line with the CLIP-style paradigm, MLLM-based embedding models typically employ the standard info-NCE\cite{info-nce} contrastive loss as their training objective. Within this framework, the query and its corresponding positive sample form a matched pair, while negative samples are generally drawn from within the same training batch. As discussed in prior works\cite{moco}\cite{simclr}\cite{LLaVE}, incorporating a greater number of hard negatives into the info-NCE loss is advantageous for model performance, as hard negatives encourage the model to learn more discriminative features rather than relying on superficial correlations.
A straightforward approach to increasing the number of hard negatives is to enlarge the training batch size. To address the associated memory constraints, SigLIP\cite{SigLIP} replaced the softmax loss in info-NCE with a sigmoid loss, rendering each term in the total loss independent and thereby reducing memory consumption to facilitate larger batch sizes. Additionally, Gao et al.\cite{GradCache} introduced a method called GradCache that decouples loss computation from embedding generation by caching gradients with respect to the embeddings. During backpropagation, gradients with respect to the model parameters are computed and accumulated across multiple chunks, enabling the use of large batch sizes.
In this work, we adopt the GradCache strategy to effectively scale up the training batch size, thereby allowing the model to benefit from a richer set of hard negatives during contrastive learning.

Beyond simply increasing the batch size, researchers have also explored explicit identification of hard negatives. These approaches can be broadly categorized into two main strategies. The first involves offline selection of hard negatives using a teacher model. For example, in \cite{sfr-embedding}\cite{NV-embed}, a pretrained teacher model is employed to mine several hard negatives for each query. In B3\cite{B3}, the authors propose clustering the entire training dataset, treating samples within the same cluster as hard negatives for one another. While offline mining significantly increases the difficulty of negative samples, the process of training a teacher model and clustering embeddings is computationally expensive, particularly for large-scale datasets. Furthermore, pre-defined hard negatives may lose their hardness as training progresses, limiting the model's ability to leverage dynamically challenging samples.
The second strategy focuses on dynamically up-weighting hard negatives during training. For instance, LLaVE\cite{LLaVE} introduces weighting factors for negative samples within the conventional info-NCE loss. An exponential weighting scheme is designed to amplify the influence of hard negatives, thereby encouraging the model to focus on more informative and challenging examples during optimization.

While these methods explicitly leverage hard negatives, the individual contribution of each hard negative sample is neither clearly quantified nor directly controlled. In this work, we begin by analyzing the gradients of the info-NCE loss with respect to the query, positive sample, and negative samples, demonstrating that the contribution of each negative sample is proportional to the probability of the query being classified as that negative. Building on this insight, we introduce a gradient modulator, termed Explicit Gradient Amplifier (EGA), which directly adjusts these probabilities to amplify the impact of hard negatives during training.
Furthermore, in contrast to LLaVE, which defines the hardness of a negative sample solely based on its similarity to the query, we argue that the relative difference between the query-negative similarity and the query-positive similarity provides a more accurate measure of negative sample hardness. The proposed EGA module modifies the conventional info-NCE loss in a computationally efficient manner and is plug-and-play compatible with any MLLM architecture.
We evaluate our approach on the Massive Multi-modal Embedding Benchmark (MMEB)\cite{VLM2Vec}. Using LLaVA-Onevision-7B\cite{llava-ov} as the backbone, our method achieves state-of-the-art performance compared to previous methods utilizing the same MLLM. Moreover, when integrated with our self-developed 7B MLLM, QQMM, our approach attains an average score of 72.5 on MMEB, securing the top position on the leaderboard.

\section{Related work}
\label{related_work}

\textbf{CLIP-series embedding models:}
CLIP\cite{CLIP} pioneered the foundational approach for multi-modal embedding by aligning image and text representations through contrastive learning. Building on this, SigLIP\cite{SigLIP} replaced the softmax-based info-NCE loss in CLIP with a sigmoid-based pairwise loss, resulting in improved embedding quality and training efficiency. Eva-CLIP\cite{Eva-CLIP} further demonstrated the scalability of this paradigm by leveraging larger datasets and model capacities to achieve more robust cross-modal representations. However, a common limitation of these CLIP-series models is that they process text and image modalities independently, which restricts their ability to reason about complex multi-modal contexts.

\textbf{MLLM-based embedding models:}
Recent advances in MLLM-based models have shown strong capabilities in generating unified representations for multi-modal inputs\cite{E5-V}\cite{VLM2Vec}\cite{UniME}\cite{LLaVE}. The introduction of the Massive Multi-modal Embedding Benchmark (MMEB) by Jiang et al.\cite{VLM2Vec} has provided a standard platform for training and evaluating such models, with VLM2Vec serving as a strong baseline trained on MMEB. LLaVE\cite{LLaVE} further improved performance by introducing a hard negative weighting strategy into the info-NCE loss, where the hardness of a negative sample is measured by its similarity to the query. UniME\cite{UniME} proposed a ranking-based hard negative mining approach, also relying on query-negative cosine similarity to determine hardness. More recently, B3\cite{B3} adopted an offline clustering method to construct batches where samples serve as hard negatives for each other. These approaches have effectively leveraged hard negatives in contrastive learning, resulting in more discriminative multi-modal embeddings.

Despite these advancements, the precise contribution of hard negatives to model updates has not been thoroughly analyzed or directly controlled. In this work, we analyze the gradient contributions of negative samples and reveal that their influence is proportional to their associated classification probabilities. Based on this insight, we propose the Explicit Gradient Amplifier (EGA), a gradient modulator that amplifies the probabilities of hard negatives during gradient computation, allowing for more direct control over their contribution. Furthermore, while previous methods typically measure the hardness of a negative sample solely by its similarity to the query, we argue that the difference between query-negative and query-positive similarities provides a more accurate assessment of hardness. Therefore, we adopt a relative similarity-based hardness score to modulate the gradient contributions of hard negatives, rather than relying exclusively on absolute query-negative similarity.

\begin{figure}
	\centering
	\includegraphics[scale=0.55]{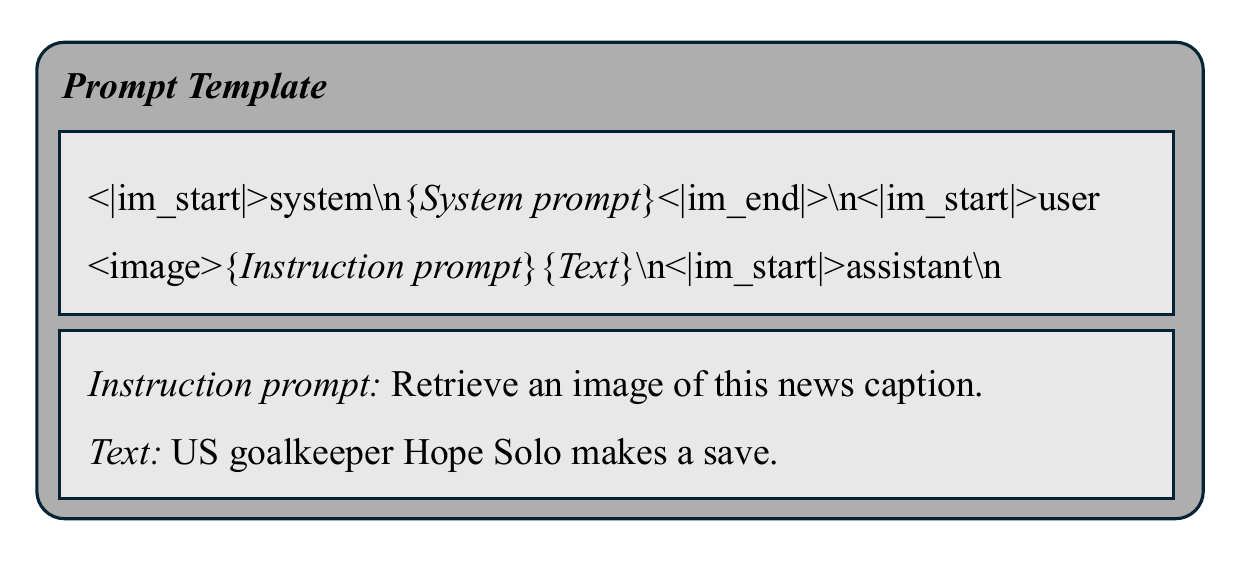}
	\caption{A template prompt for multi-modal inputs. A special <image> token is adopted to represent image input. An instruction prompt designed for each task is integerated with the image and text inputs as the full prompt following the convention adopted in VLM2Vec.}
	\label{fig: prompt}
\end{figure}

\begin{figure}
	\centering
	\includegraphics[scale=0.4]{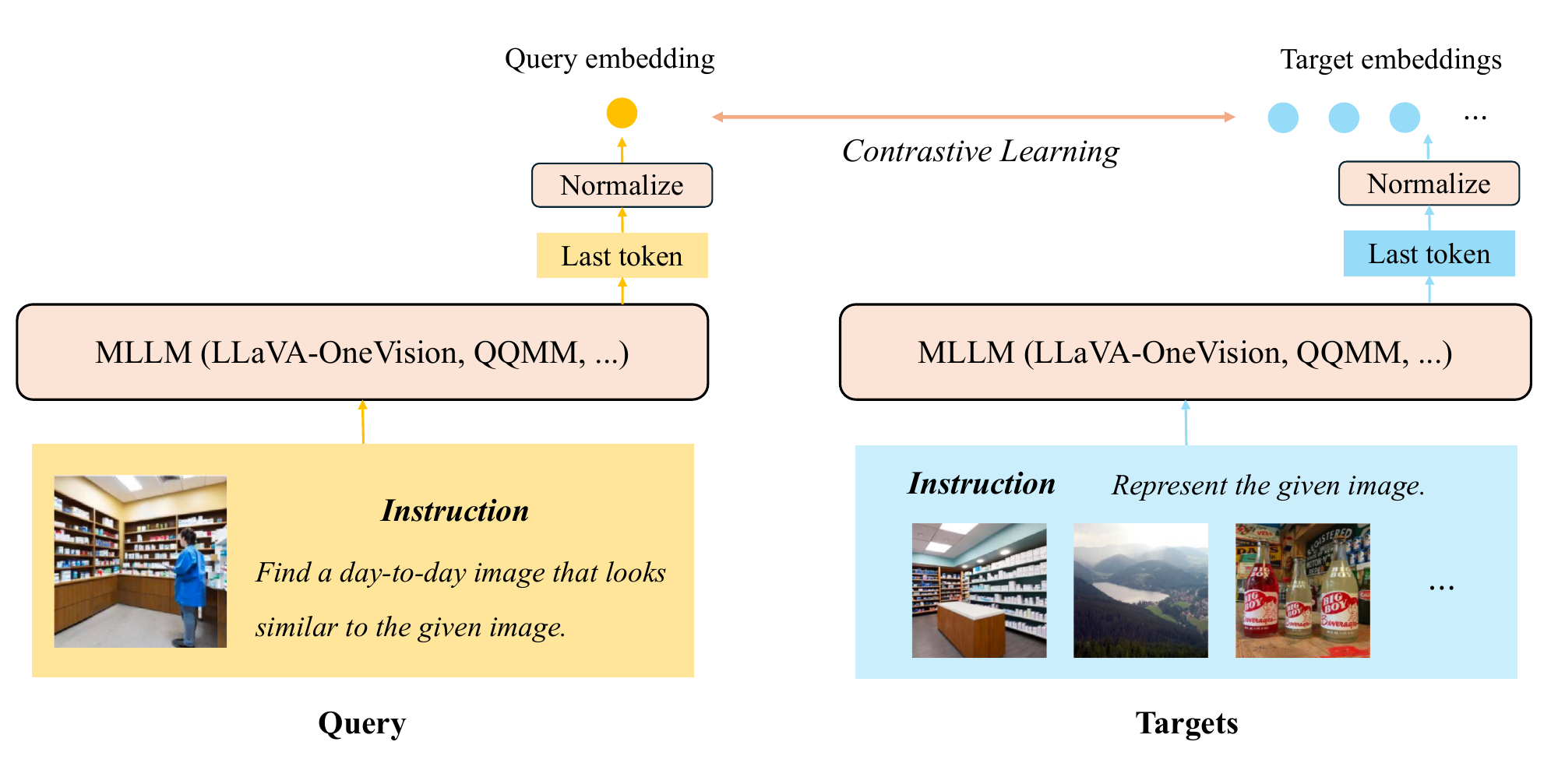}
	\caption{The overall framework of the proposed multi-modal embedding model. The MLLMs processing the query and the targets share the same architecture and parameters.}
	\label{fig: pipeline}
\end{figure}

\section{Method}
\label{method}

\subsection{Preliminary}

The info-NCE\cite{info-nce} loss serves as the foundational objective in contrastive learning. 
For each query $\mathbf{x}$, the corresponding targets consist of a positive sample $\mathbf{x^+}$ and a set of negative samples $\{\mathbf{x}^{-}_i\}$,  where $i=1, 2, ..., N-1$ .
As shown in Eq.\ref{eq:info-nce}, the loss is formulated as an $N$-class cross-entropy over the query and its $N$ targets. A temperature hyper-parameter $\tau$ is introduced to rescale the probability distribution, thereby controlling the sharpness of the softmax output.

\begin{equation}
	L = -log{\frac{e^{\mathbf{x}\cdot \mathbf{x}^+/\tau}}{e^{\mathbf{x}\cdot \mathbf{x}^+ / \tau}+ \sum_{i=1}^{N-1} e^{\mathbf{x}\cdot \mathbf{x}^{-}_{i}/\tau}}}
	\label{eq:info-nce}
\end{equation}

In CLIP-style embedding models, the query and target are typically image-text or text-image pairs. In contrast, MLLM-based multi-modal embedding models allow each query or target to contain both image and text modalities. Following the conventions in \cite{VLM2Vec}\cite{LLaVE}, we construct the input by combining the input sample with an instruction prompt, which together serve as the user prompt for embedding generation. An example of the full prompt is illustrated in Fig.~\ref{fig: prompt}.

The overall framework of our proposed method is shown in Fig.\ref{fig: pipeline}. Following the conventional practice in VLM2Vec\cite{VLM2Vec} and LLaVE\cite{LLaVE}, the constructed prompt is fed into the MLLM, and the hidden state corresponding to the last token from the final layer is extracted. This hidden state is then normalized and used as the final embedding representation. The process ensures that the embedding comprehensively captures the multi-modal semantics specified by both the sample and the instruction prompt.

\subsection{Gradient Analysis of info-NCE}

In this section, we present an analysis of the gradients associated with the info-NCE loss. By examining the gradients with respect to the query, positive sample, and negative samples, we aim to better understand how each component contributes to the optimization process and the learning dynamics of multi-modal embedding models.
Denote the gradients with respect to the query, the positive and the negative embeddings as $G_q$, $G^+$ and $G^{-}_i$ respectively.
For simplicity, we denote the N-class classification probability as $p^{+}$ and $p^{-}_i$ as shown in Eq.\ref{eq:p}.

\begin{equation}
	p^+ = \frac{e^{\mathbf{x} \cdot \mathbf{x}^+ / \tau}}{e^{\mathbf{x} \cdot \mathbf{x}^+ / \tau} + \sum_{i=1}^{N-1} e^{\mathbf{x} \cdot \mathbf{x}^{-}_i / \tau}}, \
	p^{-}_i = \frac{e^{\mathbf{x} \cdot \mathbf{x}^{-}_i / \tau}}{e^{\mathbf{x} \cdot \mathbf{x}^+ / \tau} + \sum_{i=1}^{N-1} e^{\mathbf{x} \cdot \mathbf{x}^{-}_i / \tau}}
	\label{eq:p}
\end{equation}

Eq.\ref{eq:grad_x} and Eq.\ref{eq:grad_tgt} illustrates $G_q$, $G^+$ and $G^{-}_i$.
$G_q$ can be decomposed into a weighted sum of the differences between the negative embeddings and the positive embedding. The weighting factor for each term is the probability $p^{-}_i$ that the query is classified as the $i$-th negative sample. Intuitively, the contribution of each negative sample to the update direction of the query is determined by its associated probability. Similarly, for the gradient with respect to a negative sample, $G^{-}_i$, the extent to which the negative sample is pushed away from the query is also governed by $p^{-}_i$.
Based on this gradient analysis, a straightforward approach to enhancing the impact of hard negatives is to amplify their corresponding probabilities $p^{-}_i$ during gradient computation. In the following subsection, we detail our method for identifying hard negatives and describe how their influence on training can be explicitly strengthened through the proposed Explicit Gradient Amplifier.

\begin{equation}
\begin{aligned}
	G_q = \frac{\partial L}{\partial \mathbf{x}} = \
	\frac{1}{\tau} [(p^+ - 1) * \mathbf{x}^{+} + \sum_{i=1}^{N-1} p^{-}_i * \mathbf{x}^{-}_i] = \frac{1}{\tau}\sum_{i=1}^{N-1} p^{-}_i * (\mathbf{x}^{-}_i - \mathbf{x}^+)
	\label{eq:grad_x}	
\end{aligned}
\end{equation}

\begin{equation}
		G^+ = \frac{\partial L}{\partial \mathbf{x}^{+}} = \
		\frac{1}{\tau} (p^+ - 1) * \mathbf{x}, \
		G^{-}_i = \frac{\partial L}{\partial \mathbf{x}^{-}_i} = \
		\frac{1}{\tau} p^{-}_i * \mathbf{x}
		\label{eq:grad_tgt}	
\end{equation}

\begin{figure}
	\centering
	\includegraphics[scale=0.45]{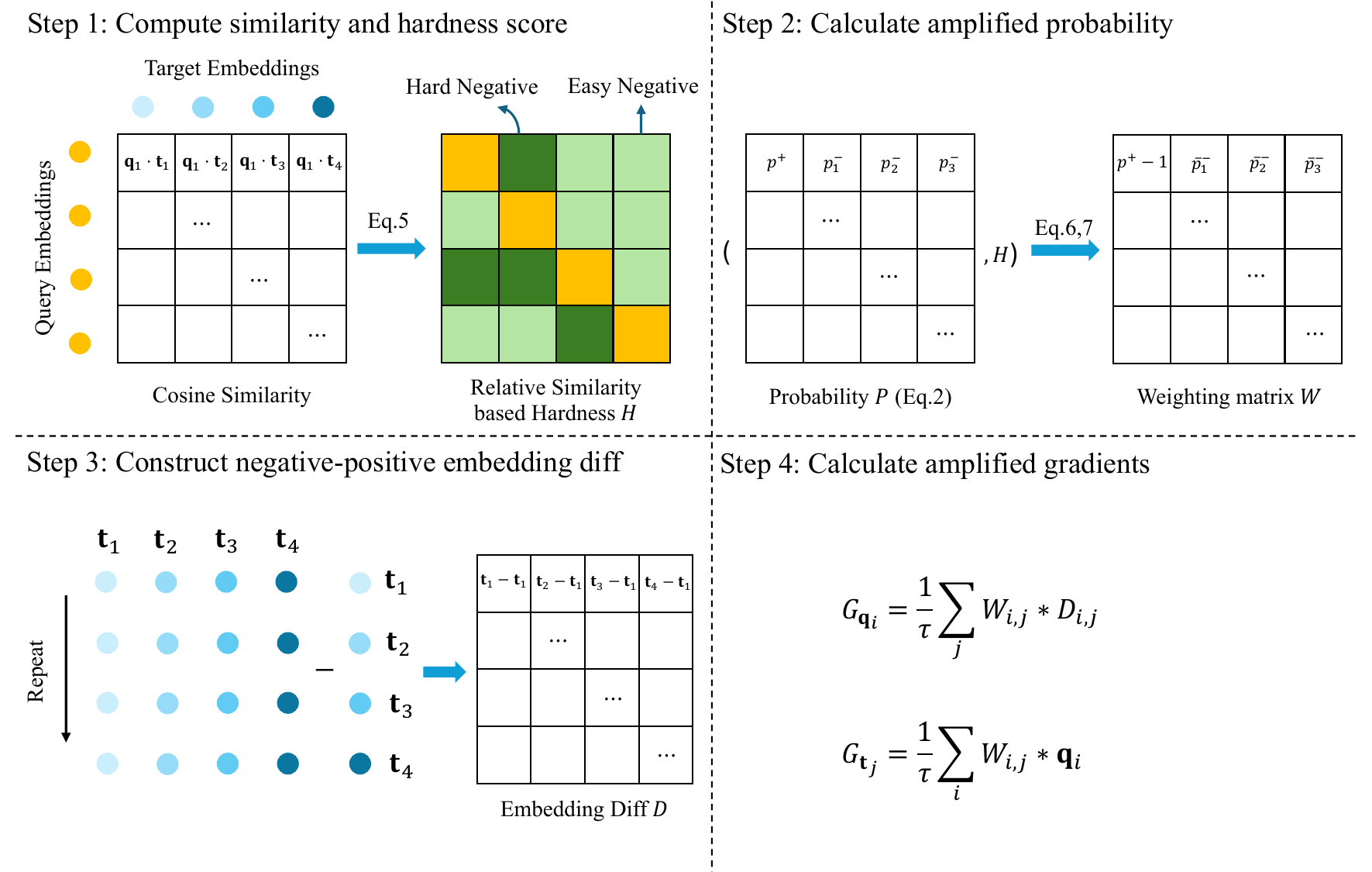}
	\caption{An overview of the process for obtaining amplified gradients in the EGA module.}
	\label{fig: EGA}
\end{figure}

\subsection{Explicit Gradient Amplifier}

Typically, a negative sample is considered hard if its similarity to the query is high. For instance, in LLaVE\cite{LLaVE}, the hardness of a negative sample is quantified by the exponential of the query-negative cosine similarity. In UniME\cite{UniME}, negative samples are ranked according to their cosine similarities with the query, and the top-$k$ negatives are selected as hard samples. While these approaches are generally effective for mining hard negatives, they overlook an important aspect: the key factor in determining whether a negative sample is hard is not its absolute similarity to the query, but rather how close its similarity is to the query compared to that of the positive sample. Specifically, if the query-positive similarity is sufficiently high (e.g., greater than 0.9), the presence of a negative sample with a moderate similarity (around 0.6) is unlikely to pose a challenge. Conversely, if the query-positive similarity is low (e.g., around 0.2), even a negative sample with a similarity of 0.3 can be considered hard. Therefore, we argue that a negative sample should be identified as hard if its similarity is close to, or even exceeds, the query-positive similarity.
To this end, for each negative sample $\mathbf{x}^{-}_i$, we compute a Relative Similarity based Hardness score (RS-H) $h^{-}_i$ as defined in Eq.\ref{eq:hardness}.
A hyper-parameter $\alpha$ is introduced to rescale the hardness distribution, allowing for flexible adjustment of the sensitivity to hard negatives.

\begin{equation}
	h^{-}_i = e^{\alpha \cdot (\mathbf{x}\cdot \mathbf{x}^{-}_i - \mathbf{x} \cdot \mathbf{x}^{+})}
	\label{eq:hardness}
\end{equation}
	
With the computed hardness scores, we can enhance the gradient contribution of hard negatives by reweighting  $p^{-}_i$ during gradient calculation.
As shown in Eq.\ref{eq:ega} and Eq.\ref{eq:ega2}, each $p^{-}_i$ is multiplied by its corresponding hardness score $h^{-}_i$ and the resulting values are then normalized across all negative samples to ensure they form a valid probability distribution.
The amplified gradients, denoted as$\overline{p}^{-}_i$ are subsequently used in the gradient computations in Eq.\ref{eq:grad_x} and Eq.\ref{eq:grad_tgt}, replacing the original $p^{-}_i$.
This reweighting mechanism allows the model to explicitly focus on hard negatives during training, thereby improving the effectiveness of contrastive learning.

\begin{equation}
	\hat{p}^{-}_i = p^{-}_{i} * h^{-}_i
	\label{eq:ega}
\end{equation}

\begin{equation}
	\overline{p}^{-}_i = \frac{\hat{p}^{-}_i}{\sum_{i=1}^{N-1} \hat{p}^{-}_i} * \sum_{i=1}^{N-1} p^{-}_i
	\label{eq:ega2}
\end{equation}

The complete Explicit Gradient Amplifier (EGA) module takes as input a batch of query embeddings $\{\mathbf{q}_i\}$ and their corresponding target embeddings $\{\mathbf{t}_i\}$, and outputs the gradients $G_\mathbf{q}$ and $G_\mathbf{t}$. For illustration, consider a case where the batch size is 4. Fig.~\ref{fig: EGA} provides a visual overview of the process for obtaining the amplified gradients. 
The procedure consists of the following steps:

\begin{itemize}
\item Step 1: Compute the similarity matrix using dot products between the query and target embeddings. The hardness matrix is then derived from these similarities according to Eq.\ref{eq:hardness}.
	
\item Step 2: Calculate the amplified probability matrix $\overline{P}$ following Eq.\ref{eq:p}, Eq.\ref{eq:ega} and Eq.\ref{eq:ega2}. Then a weighting matrix $W$ is defined as $\overline{P} - I$ where $I$  is the identity matrix.
	
\item Step 3: Construct the difference tensor $D$ between negative embeddings and positive embeddings, where for any position $(i, j)$, $D_{i, j} = \mathbf{t}_j - \mathbf{t}_i$.
	
\item Step 4: Calculate the amplified gradients. For the queries $\{\mathbf{q}_i\}$, the gradients are computed as $G_{\mathbf{q}_i} = \frac{1}{\tau}\sum_{j} W_{i, j} * D_{i,j}$.
While for the targets $\{\mathbf{t}_j\}$, the gradients are given by $G_{\mathbf{t}_j} = \frac{1}{\tau}\sum_{i} W_{i, j} * \mathbf{q}_i$.
\end{itemize}

The amplified gradient calculation depends solely on the input embeddings, making EGA a plug-and-play module compatible with any MLLM backbone. Furthermore, the entire process involves only basic tensor operations, resulting in negligible computational overhead compared to the embedding generation itself.

\begin{table}[htbp]
	\centering
	\renewcommand{\arraystretch}{1.5}
	\setlength{\tabcolsep}{4pt}
	\begin{tabular}{l|c|cccc|ccc}
		\hline
		\textbf{Models} & \textbf{\#Param} & \multicolumn{4}{c|}{\textbf{Per Task Score}} & \multicolumn{3}{c}{\textbf{Average Score}} \\ \hline
		\multicolumn{2}{c|}{} & \textbf{CLS} & \textbf{VQA} & \textbf{RET} & \textbf{GRD} & \textbf{IND} & \textbf{OOD} & \textbf{Overall} \\ \hline
		\multicolumn{2}{c|}{\textbf{\# of Datasets}} & 10 & 10 & 12 & 4 & 20 & 16 & 36 \\ \hline
		\multicolumn{9}{c}{\textit{CLIP-series embedding models}} \\ \hline
		CLIP (ViT-L)\cite{CLIP} & 0.4B & 42.8 & 9.1 & 53.0 & 51.8 & 37.1 & 38.7 & 39.2 \\ \hline
		OpenCLIP\cite{open-clip} & 0.4B & 41.5 & 6.9 & 44.6 & 53.5 & 32.8 & 36.0 & 36.6 \\ \hline
		Magiclens\cite{magiclens} & 0.4B & 38.8 & 8.3 & 35.4 & 26.0 & 31.0 & 23.7 & 27.1 \\ \hline
		SigLIP\cite{SigLIP} & 0.9B & 40.3 & 8.4 & 31.6 & 59.5 & 32.3 & 38.0 & 35.0 \\ \hline
		BLIP2\cite{blip2} & 1.2B & 27.0 & 4.2 & 33.9 & 47.0 & 25.3 & 25.1 & 28.0 \\ \hline
		CLIP (ViT-BigG/14)\cite{CLIP} & 2.5B & 52.3 & 14.0 & 50.5 & 60.3 & 38.9 & 45.8 & 44.3 \\ \hline
		EVA-CLIP\cite{Eva-CLIP} & 8B & 56.0 & 10.4 & 49.2 & 58.9 & 38.1 & 45.6 & 43.7 \\ \hline
		\multicolumn{9}{c}{\textit{MLLM-based embedding models}} \\ \hline
		E5-V (LLaVA-1.6)\cite{E5-V} & 7B & 39.7 & 10.8 & 39.4 & 60.2 & 34.2 & 33.4 & 37.5 \\ \hline
		VLM2Vec (LLaVA-OneVision)\cite{VLM2Vec} & 7B & 63.5 & 61.1 & 64.5 & 87.3 & 69.7 & 61.0 & 65.8\\ \hline
		LLaVE (LLaVA-OneVision)\cite{LLaVE}  & 7B & 65.7& 65.4 & 70.9 & 91.9 & 75.0 & 64.4 & 70.3\\ \hline
		UniME (LLaVA-OneVision)\cite{UniME} & 7B & 66.8 & 66.6 & 70.5 & 90.9 & 74.6 & 65.8 & 70.7\\ \hline
		B3 (Qwen2-VL)\cite{B3}  & 7B & 70.0 & 66.5 & 74.1 & 84.6 & 75.9 & 67.1 & 72.0\\ \hline
		\multicolumn{9}{c}{\textit{Our method}} \\ \hline
		Ours (LLaVA-OneVision) & 7B & 66.8 & 66.8 & 70.5 & 90.4 & 74.7 & 65.6 & 70.7 \\ \hline
		\textbf{Ours (QQMM)} & 7B & 69.9 & 70.0 & 72.1 & 86.0 & 77.2 & 66.6 & \textbf{72.5} \\ \hline
	\end{tabular}
	\vspace{3mm}
	\caption{Results on the MMEB benchmark. Our method achieves the state-of-the-art compared to methods using the same MLLM backbone (LLaVA-OneVision-7B). When integrated with self-developed QQMM, our method consistently outperforms previous methods.}
	\label{tab: main}
\end{table}

\section{Experiments}

\subsection{Datasets and Implementation}
\label{exp}
We conduct our experiments on the Massive Multi-modal Embedding Benchmark (MMEB), which comprises 20 datasets for training and 36 datasets for evaluation. Following established protocols in \cite{VLM2Vec}, we train our embedding model on the MMEB-train subset. To address the imbalance in data volume across different training sets, we adopt the data sampling strategy proposed in \cite{VLM2Vec}, resulting in a total of 662K training samples.

For evaluation, we assess our model on 16 out-of-distribution datasets and 20 in-distribution datasets, spanning four task categories: classification, visual question answering (VQA), retrieval, and grounding. For the classification,VQA and caption (MSCOCO\_i2t\cite{mscoco} and VisualNews\_i2t\cite{visual-news}) tasks, we further design three specific instruction prompts to guide the embedding process:

\begin{itemize}
\item \textit{Classification: "Represent the following answer to an image classification task: "}.
\item \textit{VQA: "Represent the following answer to a VQA task: "}.
\item \textit{Caption: "Represent the following answer to an image caption task: "}.
\end{itemize}

To ensure a fair comparison with prior work\cite{VLM2Vec}\cite{LLaVE}\cite{UniME}, we adopt LLaVA-OneVision-7B\cite{llava-ov} as the MLLM backbone. The model is trained for 2,000 steps with a batch size of 1,024, leveraging the GradCache framework to enable large-batch training. The hyper-parameters $\tau$ and $\alpha$ are set to 0.02 and 20.0 respectively. In addition, we report evaluation results using our self-developed QQMM model, a powerful 7B-parameter MLLM, which achieves rank-1 performance on the MMEB leaderboard.

\subsection{MMEB Evaluation}

Tab.~\ref{tab: main} presents the quantitative evaluation results of our method in comparison with previous approaches on the MMEB leaderboard. Notably, when utilizing the same MLLM backbone, LLaVA-OneVision-7B, our method consistently outperforms both VLM2Vec\cite{VLM2Vec} and LLaVE\cite{LLaVE}, with particularly clear improvements observed on out-of-distribution (OOD) datasets. Furthermore, our approach achieves performance on par with UniME\cite{UniME}, which additionally incorporates text data for distillation.
These results demonstrate the effectiveness of the proposed EGA module in enhancing the performance of contrastive learning for multi-modal embedding. Moreover, when integrated with our self-developed 7B-parameter MLLM, QQMM, our method achieves an overall score of 72.5 on MMEB, securing the top rank among all evaluated methods.

\subsection{Ablation Study}

To assess the impact of the proposed EGA module, we conducted experiments using QQMM as the backbone, following the same training protocol outlined in Section~\ref{exp}. As a baseline, we trained the model with the conventional info-NCE loss. Additionally, we evaluated a variant (Baseline + EGA) that amplifies gradients based on similarity-based hardness scores, as employed in LLaVE.
The quantitative results are summarized in Tab.~\ref{tab: ablation}. The findings indicate that amplifying the gradients associated with hard negatives leads to improved performance. Furthermore, measuring hardness based on relative similarity—rather than solely on the absolute query-negative similarity—proves to be more effective, further validating the design of our approach.

\begin{table}[htbp]
	\centering
	\renewcommand{\arraystretch}{1.5}
	\setlength{\tabcolsep}{6pt}
	\begin{tabular}{l|cccc|ccc}
		\hline
		\textbf{Models} & \multicolumn{4}{c|}{\textbf{Per Task Score}} & \multicolumn{3}{c}{\textbf{Average Score}} \\ \hline
		{} & \textbf{CLS} & \textbf{VQA} & \textbf{RET} & \textbf{GRD} & \textbf{IND} & \textbf{OOD} & \textbf{Overall} \\ \hline
		Baseline & 67.9 & 68.2 & 69.5  & 85.1 & 75.3 &  64.4 &  70.4 \\ \hline
		Baseline + EGA &  70.1 &  69.5 &  71.2 & 87.1 & 76.6 &  66.6 &  72.2\\ \hline
		Baseline + EGA + RS-H & 69.9 &  70.0 &  72.1 & 86.0 & 77.2 & 66.6 & \textbf{72.5} \\ \hline
	\end{tabular}
	\vspace{3mm}
	\caption{Ablation of the EGA module and the use of Relative Similarity based Hardness score (RS-H).}
	\label{tab: ablation}
\end{table}

\section{Conclusion}

In this work, we propose a novel gradient modulator which effectively leverages hard negatives during contrastive learning. We conduct a detailed analysis of the gradient contributions of negative samples and reveal that their influence on model updates is directly related to their associated classification probabilities. Building on this insight, we propose the Explicit Gradient Amplifier (EGA), a plug-and-play module that amplifies the gradient contributions of hard negatives by reweighting their probabilities based on a relative similarity-based hardness score. This approach enables more direct and effective control over the learning process, leading to more discriminative multi-modal embeddings.
Extensive experiments on MMEB demonstrate the effectiveness of our method. Our approach consistently outperforms previous state-of-the-art methods, particularly on out-of-distribution datasets, and achieves rank-1 performance on the MMEB leaderboard when integrated with our self-developed QQMM model. These results validate the generalizability and practical value of the proposed EGA module for advancing multi-modal embedding models.

\bibliography{ref}

\end{document}